\documentclass[letterpaper]{article} 
\usepackage{aaai2026}  
\usepackage{times}  
\usepackage{helvet}  
\usepackage{courier}  
\usepackage[hyphens]{url}  
\usepackage{graphicx} 
\usepackage{natbib}  
\usepackage{caption} 
\usepackage{algorithm}
\usepackage{algorithmic}
\usepackage{amsmath}
\usepackage{amssymb}

\usepackage{booktabs}
\usepackage{multicol}

\usepackage{multirow}
\usepackage[table,xcdraw]{xcolor}  
\usepackage{tabularx}  
\usepackage{pifont}

\newcommand{\appen}[1]{{Appendix \textcolor{red}{#1}}}
\def\ourmethod{Simple3D}
\def\ourdataset{MiniShift}

\usepackage{newfloat}
\usepackage{listings}
\DeclareCaptionStyle{ruled}{labelfont=normalfont,labelsep=colon,strut=off} 
\floatstyle{ruled}
\newfloat{listing}{tb}{lst}{}
\floatname{listing}{Listing}
\title{Towards High-Resolution 3D Anomaly Detection: A Scalable Dataset and Real-Time Framework for Subtle Industrial Defects}
\author{
    Yuqi Cheng\equalcontrib \textsuperscript{\rm 1},
    Yihan Sun\equalcontrib \textsuperscript{\rm 1},
    Hui Zhang \textsuperscript{\rm 2},
    Weiming Shen\thanks{corresponding author.} \textsuperscript{\rm 1},
    Yunkang Cao$^{\dagger}$ \textsuperscript{\rm 2}
}
\affiliations{
    \textsuperscript{\rm 1} State Key Laboratory of Intelligent Manufacturing Equipment and Technology,\\ Huazhong University of Science and Technology \\
    \textsuperscript{\rm 2}  School of Robotics, Hunan University \\
    \{yuqicheng, yihansun\}@hust.edu.cn, zhanghuihby@126.com,
    \{wshen, caoyunkang\}@ieee.org
}

\usepackage{bibentry}

\begin{document}

\maketitle

\begin{abstract}

In industrial point cloud analysis, detecting subtle anomalies demands high-resolution spatial data, yet prevailing benchmarks emphasize low-resolution inputs. To address this disparity, we propose a scalable pipeline for generating realistic and subtle 3D anomalies. Employing this pipeline, we developed \textbf{MiniShift}, the inaugural high-resolution 3D anomaly detection dataset, encompassing 2,577 point clouds, each with 500,000 points and anomalies occupying less than 1\% of the total. We further introduce \textbf{Simple3D}, an efficient framework integrating Multi-scale Neighborhood Descriptors (MSND) and Local Feature Spatial Aggregation (LFSA) to capture intricate geometric details with minimal computational overhead, achieving real-time inference exceeding 20 fps. Extensive evaluations on \textbf{MiniShift} and established benchmarks demonstrate that \textbf{Simple3D} surpasses state-of-the-art methods in both accuracy and speed, highlighting the pivotal role of high-resolution data and effective feature aggregation in advancing practical 3D anomaly detection.

\end{abstract}
%


\begin{links}
    \link{HomePage}{https://hustcyq.github.io/MiniShift-Simple3D}
    \link{Code}{https://github.com/hustCYQ/MiniShift-Simple3D}
    \link{Datasets}{https://huggingface.co/datasets/ChengYuQi99/MiniShift}
\end{links}

\section{Introduction}

\begin{figure*}[t!]
\centering\includegraphics[width=\linewidth]{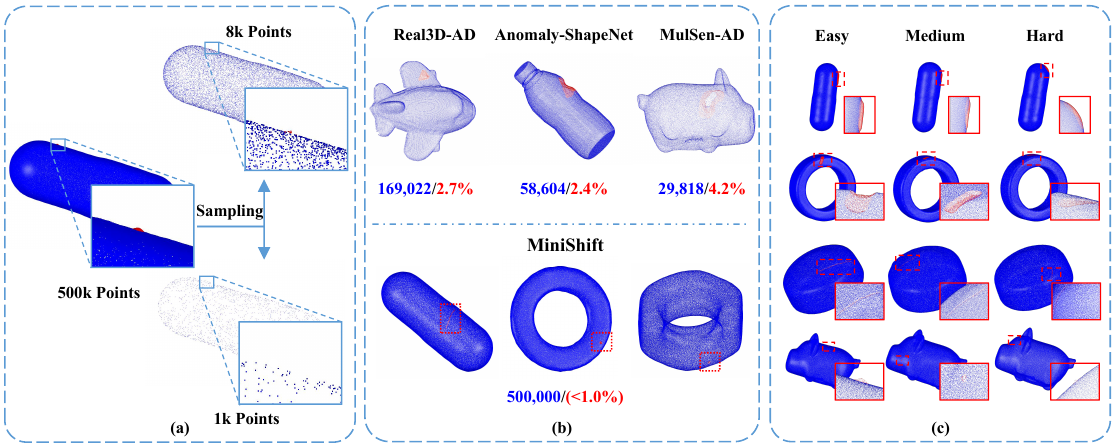}
\caption{
(a) Detecting subtle defects requires high-resolution point clouds to ensure precise localization and accurate identification. 
(b) The proposed \textbf{MiniShift} dataset features high-density point clouds with approximately 500k points per sample, where anomalies occupy less than 1\% of the total surface area—posing a significantly more challenging scenario than existing benchmarks. 
\textcolor{blue}{Blue numbers} indicate the average number of points per sample, while \textcolor{red}{red numbers} denote the anomaly occupancy rate. 
(c) Visualization of representative samples from MiniShift. Each row corresponds to a different defect category: {Areal}, {Striate}, {Scratch}, and {Sphere}.
}
\vspace{-3mm}
\label{fig:teaser}
\end{figure*}

Despite the exceptional precision of modern manufacturing processes, subtle imperfections often persist, eluding conventional inspection techniques and introducing significant safety risks in downstream applications~\cite{MVP-PCLIP}. Unsupervised point cloud anomaly detection emerges as a compelling strategy for identifying these hard-to-detect flaws~\cite{AdaCLIP,M3DM,PO3AD}. However, the efficacy of such methods is curtailed by their reliance on the widely utilized low-resolution evaluation framework. This dependency creates a pronounced gap between cutting-edge research and the rigorous demands of industrial implementation.

Conventionally, prevalent approaches for 3D anomaly detection begin by selecting a sparse subset of points (typically around 1k) from the original point cloud. These methods encode local neighborhood features to form point groups and assign an anomaly score to each group. Reconstruction-based techniques, such as IMRNet~\citep{shape_anomaly} and R3D-AD~\citep{r3d}, transform each group into representative tokens, which are then reconstructed via a point transformer architecture. The anomaly score is derived by computing the feature-wise discrepancy between the original and reconstructed tokens. In contrast, prototype-based methods like ISMP~\citep{LookInside} and GLFM~\citep{GLFM} utilize a pre-trained Point Transformer to extract discriminative features from the point groups, and quantify anomalies based on their deviation from learned prototype representations.
To generate dense anomaly maps, these sparse group-level scores are interpolated across the full point cloud. However, such interpolation inevitably compromises the spatial granularity necessary for detecting fine-grained defects. For example, an imperfection measuring just 1mm $\times$ 1mm on a 20mm $\times$ 20mm surface becomes almost imperceptible when the point cloud is reduced to 1k points. As illustrated in Figure~\ref{fig:teaser} (a), a subtle protrusion is clearly discernible in the full-resolution 500k point cloud and remains detectable at 8k points, yet it vanishes almost entirely after downsampling to 1k points. This observation highlights the critical need for high-resolution anomaly detection frameworks that preserve spatial fidelity to effectively identify subtle industrial defects.

To detect such subtle anomalies, industrial inspection systems often capture hundreds of thousands to over a million points per object~\cite{MVGR}. Although datasets like MVTec 3D-AD\cite{mvtec3d}, Real3D-AD\cite{read3d}, Anomaly-ShapeNet\cite{shape_anomaly}, and MulSenAD~\cite{MulSen-AD} offer high-resolution point clouds, their anomalies tend to be relatively prominent, contrasting sharply with the exceedingly subtle irregularities encountered in real-world contexts. To bridge this divide between public datasets and practical applications, we propose an innovative pipeline for synthesizing realistic, subtle anomalies within high-resolution point clouds. Additionally, to systematically evaluate detection performance across varying scales, we introduce a multi-level difficulty protocol, classifying anomalies into three tiers—easy, medium, and hard—based on their geometric perturbations and visual detectability. Applying this pipeline to a subset of normal samples from MulSenAD~\cite{MulSen-AD}, we establish a novel high-resolution benchmark, \textbf{MiniShift}. As illustrated in Figure~\ref{fig:teaser} (b), MiniShift features point clouds with 500k points, far surpassing the resolution of existing datasets (typically below 170k points). Crucially, its anomalies constitute less than 1\% of the surface area, necessitating the use of high-resolution spatial features and posing a markedly greater challenge than prior benchmarks.

We conducted an extensive evaluation of state-of-the-art (SOTA) methods on MiniShift using point clouds with 8k points, revealing that these approaches fall short in both performance and efficiency. 
This inadequacy stems primarily from the limitations of CNN-~\cite{ResNet} and ViT-~\cite{PT} based backbones in applying high-resolution point cloud detection: (i) prohibitive computational complexity that scales with input group quantity, and (ii) inaccurate representation of local geometric information. In response, we revisit the utility of handcrafted point cloud descriptors and present a streamlined yet robust baseline, \textbf{Simple3D}, tailored for high-resolution anomaly detection. Simple3D integrates two novel components: the Multi-scale Neighborhood Descriptor (MSND) and Local Feature Spatial Aggregation (LFSA). By computing multi-scale descriptors for each point and hierarchically aggregating local spatial features through progressive downsampling, Simple3D achieves a detailed characterization of local geometry. Leveraging lightweight descriptors, it delivers substantially improved accuracy and efficiency over existing methods. Comprehensive experiments affirm that Simple3D attains SOTA performance not only on MiniShift but also across traditional benchmarks, while sustaining real-time inference speeds exceeding 20 FPS.
In summary, our contributions are as follows:

\begin{itemize}
\item We present MiniShift, a high-resolution 3D anomaly detection dataset comprising 2,577 point clouds, each with 500k points, accompanied by a triple-level difficulty framework to thoroughly assess detection methods.
\item We introduce Simple3D, a high-resolution anomaly detection approach with two core components—MSND and LFSA—that markedly enhances the identification of subtle geometric anomalies while preserving computational efficiency.
\item Through rigorous experimentation, we demonstrate that Simple3D consistently surpasses existing SOTA methods across MiniShift, and other existing datasets including Real3D-AD, Anomaly-ShapeNet, and MulSenAD, achieving superior accuracy and real-time performance.
\item Our systematic analysis underscores the critical role of high-resolution input in precise 3D anomaly detection, revealing that judiciously aggregated local geometric features serve as effective representations for this purpose.
\end{itemize}

\section{MiniShift Dataset}
\subsection{Pipeline}
\begin{figure}[t]
\centering\includegraphics[width=\linewidth]{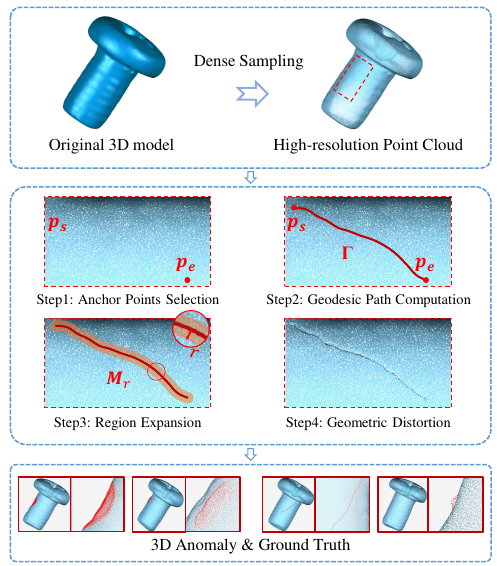}
\caption{\textbf{The construction pipeline of MiniShift}. The 3D models are utilized to generate various defects through the proposed AG-GAS framework.}
\vspace{-3mm}
\label{fig:dataset_pipeline}
\end{figure}

The development of the MiniShift dataset is illustrated in Figure~\ref{fig:dataset_pipeline}. We commence by extracting 3D data from 12 distinct categories of typical industrial components within the MulSen-AD dataset. These data undergo dense sampling to produce high-resolution point clouds. Subsequently, we propose the Anchor-Guided Geometric Anomaly Synthesis (AG-GAS) method, which facilitates the creation of four anomaly types—namely Areal, Striate, Scratch, and Sphere—alongside their associated defect masks. This synthesis is achieved through precise manipulation of anchor point positions and geometric deformation parameters. Additionally, we implement a difficulty protocol, categorizing the dataset into three subsets—easy, medium, and hard—to replicate real-world industrial scenarios characterized by varying detection complexities. Comprehensive details regarding MiniShift, encompassing categories, sample sizes for training and testing sets, and anomaly distributions, are documented in \appen{A}.

\subsection{Anchor-Guided Geometric Anomaly Synthesis}
We introduce an automated framework engineered to synthesize realistic and diverse 3D defects, as depicted in Figure~\ref{fig:dataset_pipeline}. The process begins with the random selection of two anchor points, followed by the determination of connecting path points. These path points are then expanded into a local neighborhood, delineating the deformation region wherein a stretching operation generates the defect. Unlike anomaly synthesis methods in R3D~\cite{r3d} and GLFM~\cite{GLFM}, which predominantly simulate basic protrusions or depressions within localized circular zones, our framework affords the flexibility to craft defects of diverse shapes and scales through minimal parameter adjustments. This adaptability enhances the simulation of the randomness and intricacy characteristic of industrial defects. The framework is organized into four sequential stages:

\noindent\textbf{Anchor Points Selection:} 
Two anchor points, denoted $\boldsymbol{p}_s$ and $\boldsymbol{p}_e \in \boldsymbol{P}$, are selected from the input point cloud $\boldsymbol{P} \in \mathbb{R}^{n \times 3}$, comprising $n$ points, to regulate the position and span of the synthetic anomaly.

\noindent\textbf{Geodesic Path Computation:} 
As illustrated in Figure~\ref{fig:dataset_pipeline}, we calculate the geodesic path points between the anchor points to support subsequent anomaly generation. Initially, we construct an undirected graph $G = (\{\boldsymbol{p}_i\}, \{\boldsymbol{e}_{ij}\})$ by defining edges $\boldsymbol{e}_{ij} = \langle \boldsymbol{p}_i, \boldsymbol{p}_j \rangle$, where $\boldsymbol{p}_i \in \boldsymbol{P}$ and $\boldsymbol{p}_j$ represents one of the $k$ nearest neighbors of $\boldsymbol{p}_i$ within $\boldsymbol{P}$. Dijkstra’s algorithm~\cite{Dijkstra} is then applied to ascertain the shortest path $\pi^*$ between $\boldsymbol{p}_s$ and $\boldsymbol{p}_e$, yielding the geodesic path point set $\boldsymbol{\Gamma}$:
\begin{equation}
\pi^* = \mathop{\arg\min}\limits_{\pi \in \Pi(\boldsymbol{p}_s, \boldsymbol{p}_e)} \sum_{\boldsymbol{e}_{ij} \in \pi} w(\boldsymbol{e}_{ij})
\end{equation}
\begin{equation}
\boldsymbol{\Gamma} = \{\{\boldsymbol{p}_i, \boldsymbol{p}_j\} \mid \boldsymbol{e}_{ij} \in \pi^*\}
\end{equation}
where $\Pi(\boldsymbol{p}_s, \boldsymbol{p}_e)$ denotes all possible paths from $\boldsymbol{p}_s$ to $\boldsymbol{p}_e$, $\pi$ is a path within this set, and $w(\boldsymbol{e}_{ij})$ is the Euclidean distance between $\boldsymbol{p}_i$ and $\boldsymbol{p}_j$.

\noindent\textbf{Region Expansion:} 
Utilizing the geodesic path $\boldsymbol{\Gamma}$ as the central axis, we define the mask region $\boldsymbol{M}_r$ through expansion with a control radius $r$, encompassing points in $\boldsymbol{P}$ that contribute to anomaly generation:
\begin{equation}
\boldsymbol{M}_r = \left\{ \boldsymbol{p}_j \in \boldsymbol{P} \,\big|\, \|\boldsymbol{p}_j - \boldsymbol{p}_i\| < r, \forall \boldsymbol{p}_i \in \boldsymbol{\Gamma} \right\}
\end{equation}

\noindent\textbf{Geometric Distortion:} 
The distortion direction is established by computing the average normal vector $\bar{\boldsymbol{n}}_{avg}$ of points within $\boldsymbol{M}_r$:
\begin{equation}
\bar{\boldsymbol{n}}_{avg} = \frac{\bar{\boldsymbol{n}}}{\|\bar{\boldsymbol{n}}\|_2}, \quad \bar{\boldsymbol{n}} = \frac{1}{|\boldsymbol{M}_r|} \sum_{\boldsymbol{p}_j \in \boldsymbol{M}_r} \boldsymbol{n}_j
\end{equation}
where $\boldsymbol{n}_j$ is the normal vector of point $\boldsymbol{p}_j$, and $|\boldsymbol{M}_r|$ indicates the number of points in $\boldsymbol{M}_r$. The distortion, controlled by a parameter $dir \in \{1, -1\}$ (1 for protrusion, -1 for depression), shifts points in $\boldsymbol{M}_r$ by a distance $d$ along $\bar{\boldsymbol{n}}_{avg}$. To ensure smoothness, the stretching distance is scaled according to proximity to the central axis, yielding the anomaly point:
\begin{equation}
\boldsymbol{p}_j^{\prime} = \boldsymbol{p}_j + dir \cdot \bar{\boldsymbol{n}}_{avg} \cdot \left(1 - \frac{\boldsymbol{d}_j}{\boldsymbol{d}_{\text{max}}}\right) \cdot d, \quad \boldsymbol{d}_{\text{max}} = \max_{\boldsymbol{p}_i \in \boldsymbol{M}_r} \boldsymbol{d}_i
\end{equation}
where $\boldsymbol{d}_j = \mathop{\min}\limits_{\boldsymbol{p}_m \in \boldsymbol{\Gamma}} \|\boldsymbol{p}_j - \boldsymbol{p}_m\|$.

Through precise parameterization, our framework synthesizes diverse defect types for the MiniShift dataset, selecting four representative categories—Areal, Striate, Scratch, and Sphere—reflecting prevalent industrial defect patterns. Areal and Striate defects, marked by subtle distortions, commonly result from uneven mechanical stress or thermal gradients in machining and casting. Conversely, Scratch and Sphere anomalies, with minimal surface coverage, typically arise from impacts or friction during milling, handling, or assembly. We generate 30 unique samples per category and defect type. Further dataset specifics are provided in \appen{A}, with comparisons to other datasets in \appen{B}.

\subsection{Difficulty Protocol}
To tackle the dual objectives of industrial inspection realism and algorithmic performance evaluation, we devise a difficulty protocol that classifies synthesized anomalies into three tiers—easy, medium, and hard—based on geometric saliency and perceptual visibility. Large-scale deformations and expansive defects are readily detectable by human inspectors and automated systems, whereas subtle anomalies (e.g., minor protrusions or micro-scratches) exhibit higher oversight rates. The easy subset features anomalies with prominent distortions or broad spatial extent, facilitating identification, while the hard subset includes minute or nearly imperceptible defects, mirroring critical industrial edge cases prone to missed detection. Figure~\ref{fig:teaser} (c) showcases anomaly samples across difficulty levels, with additional visualizations in \appen{G}. This protocol augments the dataset’s hierarchical representation of real-world defect complexity and establishes a benchmark for assessing algorithmic robustness and generalization.

The parameters governing anomaly difficulty are outlined in Table~\ref{table:Protocol}, adjusted via the ranges of $\alpha$, $\beta$, and $\gamma$, defined as:
\begin{equation}
\alpha = l/D, \quad \beta = r/D, \quad \gamma = d/D,
\end{equation}
where $D = \|\max(\boldsymbol{P}) - \min(\boldsymbol{P})\|_2$ represents the point cloud’s bounding box diagonal, $l$ is the geodesic distance between $\boldsymbol{p}_s$ and $\boldsymbol{p}_e$.

\begin{table}[h]
\centering
\caption{\textbf{Difficulty Protocol}. Color-filled cells highlight the key indicators that determine the difficulty of the category.}
\label{table:Protocol}
\fontsize{11}{14}\selectfont{
\resizebox{\linewidth}{!}{
\begin{tabular}{cc|c|c|c}
\toprule[1.5pt]
\multicolumn{2}{c|}{Protocol}  & Easy &  Medium & Hard \\  \midrule

\multirow{3}{*}{Areal}   & $\alpha$ & $0.05\sim0.1$ & $0.05\sim0.1$ & $0.05\sim0.1$ \\ 
   & $\beta$ & $0.04\sim0.07$ & $0.04\sim0.07$ & $0.04\sim0.07$ \\
   & \cellcolor{blue!8}{${\gamma}$} & \cellcolor{blue!8}{$5\times10^{-3}\sim7\times10^{-3}$} & \cellcolor{blue!8}{$3\times10^{-3}\sim5\times10^{-3}$} & \cellcolor{blue!8}{$1\times10^{-3}\sim3\times10^{-3}$} \\ \hline

\multirow{3}{*}{Striate}   & $\alpha$ & $0.02\sim0.1$ & $0.02\sim0.1$ & $0.02\sim0.1$ \\ 
   & $\beta$ & $0.01\sim0.03$ & $0.01\sim0.03$ & $0.01\sim0.03$ \\
   & \cellcolor{blue!8}{$\gamma$} & \cellcolor{blue!8}{$5\times10^{-3}\sim7\times10^{-3}$} & \cellcolor{blue!8}{$3\times10^{-3}\sim5\times10^{-3}$} & \cellcolor{blue!8}{$1\times10^{-3}\sim3\times10^{-3}$} \\ \hline

\multirow{3}{*}{Scratch}   & \cellcolor{blue!8}{$\alpha$} & \cellcolor{blue!8}{$0.3\sim0.4$} & \cellcolor{blue!8}{$0.2\sim0.3$} & \cellcolor{blue!8}{$0.1\sim0.2$} \\ 
   & \cellcolor{blue!8}{$\beta$} & \cellcolor{blue!8}{$3\times10^{-3}\sim4\times10^{-3}$} & \cellcolor{blue!8}{$2\times10^{-3}\sim3\times10^{-3}$} & \cellcolor{blue!8}{$1\times10^{-3}\sim2\times10^{-3}$} \\
   & $\gamma$ & $1\times10^{-3}\sim5\times10^{-3}$ & $1\times10^{-3}\sim5\times10^{-3}$ & $1\times10^{-3}\sim5\times10^{-3}$ \\ \hline

\multirow{3}{*}{Sphere}   & $\alpha$ & $<1\times10^{-3}$ & $<1\times10^{-3}$ & $<1\times10^{-3}$ \\ 
   & \cellcolor{blue!8}{$\beta$} & \cellcolor{blue!8}{$7\times10^{-3}\sim9\times10^{-3}$} & \cellcolor{blue!8}{$5\times10^{-3}\sim7\times10^{-3}$} & \cellcolor{blue!8}{$3\times10^{-3}\sim5\times10^{-3}$} \\
   & $\gamma$ & $1\times10^{-3}\sim5\times10^{-3}$ & $1\times10^{-3}\sim5\times10^{-3}$ & $1\times10^{-3}\sim5\times10^{-3}$ \\

 \bottomrule[1.5pt]

\end{tabular}
}
}
\vspace{-3mm}
\end{table}
\begin{figure*}[t!]
\centering
\includegraphics[width=\linewidth]{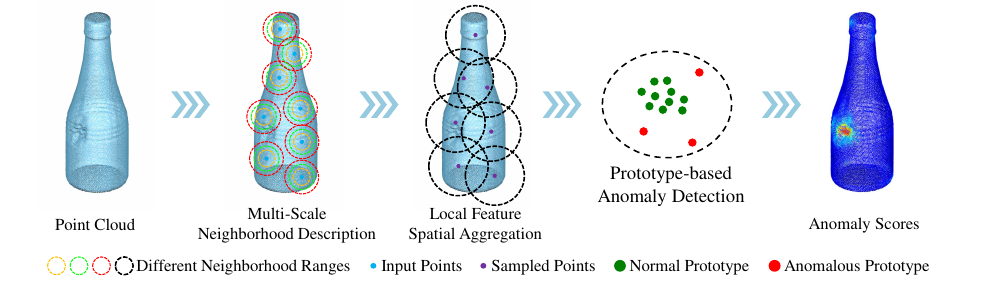}
\caption{\textbf{Framework of the proposed method, \ourmethod{}.} High-resolution point clouds undergo initial processing to derive multi-scale neighborhood descriptors. These descriptors are then spatially aggregated to yield enhanced point cloud features. \ourmethod{} utilizes a prototype-based approach to facilitate both object-wise and point-wise anomaly detection.}
\vspace{-3mm}
\label{fig:method_framework}
\end{figure*}

\section{Simple3D Method}

\subsection{Overview}

To address the challenges posed by high-resolution detection, we present \ourmethod{}, a simple yet effective framework. The overall architecture of this framework is illustrated in Figure~\ref{fig:method_framework}. Unlike existing approaches that commonly utilize CNNs or ViTs for point cloud feature extraction, resulting in prohibitive computational costs for high-resolution detection, our method focuses on extracting local point features using handcrafted descriptors. These features are subsequently aggregated across spatial hierarchies to enhance the detection of anomalies. \ourmethod{} employs a prototype-based detection mechanism, where anomaly scores are assigned by quantifying the deviations of features from established normal prototypes.

\subsection{Multi-Scale Neighborhood Description (MSND)}

Let the input point cloud be denoted by $\boldsymbol{P} \in \mathbb{R}^{n \times 3}$, where $n$ represents the total number of points. For each point $\boldsymbol{p}_i \in \boldsymbol{P}$, we identify its $k$ nearest neighbors to construct a point set $\boldsymbol{R}_i$. To harness the complementary information inherent in local point distributions across various neighborhood scales, we employ multiple neighbor counts $k_1, k_2, \dots, k_m$, thereby generating a sequence of point sets $\boldsymbol{R}_{i1}, \boldsymbol{R}_{i2}, \dots, \boldsymbol{R}_{im}$. Subsequently, a local feature operator $\boldsymbol{f}$ is applied to extract feature descriptors from each point set. These descriptors are concatenated to form the multi-scale local feature representation $\boldsymbol{F}_i$ of $\boldsymbol{p}_i$:

\begin{equation}
\boldsymbol{F}_i = \text{Concat}(\boldsymbol{R}_{i1}, \boldsymbol{R}_{i2}, \dots, \boldsymbol{R}_{im})
\end{equation}

\subsection{Local Feature Spatial Aggregation (LFSA)}

To further refine the feature representation, we randomly sample $t$ points from $\boldsymbol{P}$ and aggregate the MSND within their respective neighborhoods. This process yields enhanced point features that possess both an expanded receptive field and enriched local geometric information. Specifically, let the sampled points be denoted as $\boldsymbol{p}_{s_j}$ for $0 \leq j \leq t$ and $0 \leq s_j \leq n$, with the neighbor count specified as $k_L$. The enhanced point feature $\boldsymbol{F}^A_{s_j}$ for $\boldsymbol{p}_{s_j}$ is computed as:

\begin{equation}
\boldsymbol{F}^A_{s_j} = \frac{1}{k_L} \sum_{\boldsymbol{F} \in \boldsymbol{R}_{s_j k_L}} \boldsymbol{F}
\end{equation}

\subsection{Anomaly Detection}

The enhanced point features derived from normal point clouds are utilized to establish a normal prototype set $\boldsymbol{\mathcal{S}} = \{\boldsymbol{F}^A_{s_1}, \boldsymbol{F}^A_{s_2}, \ldots, \boldsymbol{F}^A_{s_t}\}$. For the enhanced point features $\boldsymbol{F}^A_{\text{test}}$ of test point clouds, deviations from this normal prototype set are interpreted as anomaly scores. The point-wise anomaly scores $\boldsymbol{A}$ and the object-wise anomaly score $\xi$ are determined by:

\begin{equation}
\boldsymbol{A} = \|\boldsymbol{F}^A_{\text{test}} - \boldsymbol{F}^{*}\|
\end{equation}

\begin{equation}
\xi = \max(\boldsymbol{A})
\end{equation}

\noindent where $\boldsymbol{F}^{*} =  \mathop{\min}\limits_{s \in \boldsymbol{\mathcal{S}}} \|s - \boldsymbol{F}^A_{\text{test}}\|$.

\section{Experiment}

\subsection{Experimental Setups}

\noindent\textbf{Datasets:}
Our experimental evaluation encompasses the proposed \textbf{MiniShift} dataset alongside three established public benchmarks: \textbf{Real3D-AD}~\cite{read3d}, \textbf{Anomaly-ShapeNet}~\cite{shape_anomaly}, and \textbf{MulSen-AD}~\cite{MulSen-AD}. We refer you to \appen{B} for more details about these datasets.

\noindent\textbf{Implementation Details:}
For point cloud feature extraction, our approach utilizes the handcrafted descriptor FPFH~\cite{FPFH} due to its efficient CUDA-accelerated implementation available in Open3D~\cite{Open3D}. The multi-scale ranges are set to 40, 80, and 120 by default, and the number of aggregation points is configured to 128. All experiments are conducted on NVIDIA A100 GPUs.

\noindent\textbf{Evaluation Metrics:}
To assess the efficacy of our method, we employ the widely recognized Area Under the Receiver Operating Characteristic curve (AUROC) metric. Specifically, we calculate both object-wise AUROC (O-ROC) and point-wise AUROC (P-ROC) to gauge performance across the four datasets.

\noindent\textbf{Comparison Methods:}
Our proposed method is rigorously benchmarked against a diverse array of recent SOTA approaches. These include prototype-based methods such as PatchCore~\cite{PatchCore,BTF}, BTF~\cite{BTF}, M3DM~\cite{M3DM}, CPMF~\cite{cpmf}, Reg3D-AD~\cite{read3d}, Group3AD~\cite{Group3d}, and GLFM~\cite{GLFM}; reconstruction-based methods like R3D-AD~\cite{r3d}, IMRNet~\cite{shape_anomaly}, and MC3D-AD~\cite{MC3D-AD}; and the regression-based method PO3AD~\cite{PO3AD}. Notably, M3DM and PatchCore are evaluated with various feature types, denoted as ``-FP'' for FPFH~\cite{FPFH}, ``-PM'' for PointMAE~\cite{pointmae}, ``-PB'' for Point-BERT~\cite{PointBert}.

\begin{table}[t]
\centering
\caption{\textbf{Quantitative Results on \ourdataset}. The results are presented in O-ROC\%/P-ROC\%. The best performance is in \textbf{bold}, and the second best is \underline{underlined}.}
\label{table:minishift}
\fontsize{11}{14}\selectfont{
\resizebox{\linewidth}{!}{
\begin{tabular}{c|ccccc|
>{\columncolor{blue!8}}c}
\toprule[1.5pt]

Method~$\rightarrow$  &  \small{PatchCore-FP} &  \small{PatchCore-PM} &  \small{PatchCore-PB}  & \small{R3D-AD}  &  \small{GLFM}   &  \textbf{\ourmethod}  \\
Level~$\downarrow$   & \small{CVPR'22} &  \small{CVPR'22} & \small{CVPR'22} & \small{ECCV'24} &  \small{TASE'25} &   \textbf{Ours}  \\ \midrule

Easy   & \underline{68.3}/56.5 & 57.2/57.3 & 55.5/51.6 & 56.3/48.3 & 57.7/\underline{67.7} & \textbf{75.6}/\textbf{77.3} \\ \hline
Medium & \underline{65.6}/54.1 & 55.8/52.6 & 46.6/50.3 & 50.8/50.7 & 55.0/\underline{57.3} & \textbf{68.6}/\textbf{65.5} \\ \hline
Hard   & \underline{61.4}/51.1 & 54.3/52.0 & 52.9/50.8 & 50.1/50.6 & 52.8/\underline{52.7} & \textbf{61.6}/\textbf{56.3} \\ \hline
ALL    & \underline{65.1}/53.7 & 55.9/54.1 & 50.0/51.4 & 53.6/49.8 & 55.8/\underline{58.7} & \textbf{68.6}/\textbf{66.2} \\

\bottomrule[1.5pt]

\end{tabular}}}
\end{table}

\begin{figure}[h]
\centering\includegraphics[width=\linewidth]{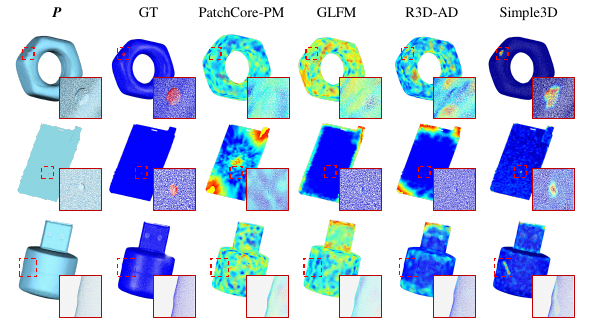}
\caption{\textbf{Visualization of our point-wise anomaly detection results on \ourdataset{}.} From top to bottom: input point clouds, ground truths, anomaly maps.}
\vspace{-3mm}
\label{fig:vis_minishift}
\end{figure}

\begin{table*}[t]
\centering
\caption{\textbf{Quantitative Results on Real3D-AD, Anomaly-ShapeNet and MulSen-AD}. The results are presented in O-ROC\%/P-ROC\%. The best performance is in \textbf{bold}, and the second best is \underline{underlined}.}
\label{table:public_results}
\fontsize{11}{14}\selectfont{
\resizebox{\linewidth}{!}{
\begin{tabular}{c|cccccccc|
>{\columncolor{blue!8}}c}
\toprule[1.5pt]

Method~$\rightarrow$   & \small{CPMF} & 
\small{Reg3D-AD} & \small{Group3AD}   & \small{IMRNet}   & \small{ISMP}  & \small{GLFM}    &  \small{PO3AD}  &  \small{MC3D-AD}   & \textbf{Simple3D}    \\
Real3D-AD  & \small{PR'24} & \small{NeurIPS'23} & \small{ACM MM'24}   & \small{CVPR'24} & \small{AAAI'25} & \small{TASE'25} & \small{CVPR'25} & \small{IJCAI'25} & \textbf{Ours}   \\ \midrule

O-ROC/P-ROC  & 62.5/75.9 & 70.4/70.5 & 75.1/73.5 & 72.5/- & 76.7/\underline{83.6} & 75.0/76.7 & 76.5/- & \underline{78.2}/76.8 & \textbf{80.4}/\textbf{92.3}   \\ \bottomrule[1.5pt]

Method~$\rightarrow$  & \small{M3DM-PM}  & \small{CPMF} & 
\small{Reg3D-AD}  & \small{IMRNet}   & \small{ISMP}   &  \small{GLFM}  &  \small{PO3AD}  &  \small{MC3D-AD}   & \textbf{Simple3D}     \\
Anomaly-ShapeNet  & \small{CVPR'23}  & \small{PR'24} & \small{NeurIPS'23} & \small{CVPR'24}   & \small{AAAI'25}  &  \small{TASE'25} & \small{CVPR'25} & \small{IJCAI'25} & \textbf{Ours}   \\ \midrule

O-ROC/P-ROC        & 51.1/54.9 & 55.9/- & 57.2/- & 66.1/65.0 & 75.7/69.1 & 61.9/74.5 & 83.9/\underline{89.8}  & \underline{84.2}/74.8  & \textbf{86.0}/\textbf{92.9}     \\ \bottomrule[1.5pt]

Method~$\rightarrow$  & \small{M3DM-PM} & \small{M3DM-PB} & \small{PatchCore-FP} &\small{PatchCore-FP-R} &  \small{PatchCore-PM} &  \small{IMRNet}   & \small{Reg3D-AD} & GLFM   &\textbf{Simple3D}    \\
MulSen-AD & \small{CVPR'23}  & \small{CVPR'23} & \small{CVPR'22} & \small{CVPR'22}   & \small{CVPR'22} & \small{CVPR'24} & \small{NeurIPS'23} & \small{TASE'25}   & \textbf{Ours}   \\ \midrule

O-ROC/P-ROC     & 62.8/58.7 & 70.5/61.1 & \underline{86.0}/64.0     & 83.3/62.0   & 84.0/60.5   & 60.1/46.7 & 74.9/64.1 & 78.5/66.5 & \textbf{88.2}/\textbf{80.3}
 \\ 
\bottomrule[1.5pt]

\end{tabular}}}
\end{table*}

\begin{figure*}[t!]
\centering\includegraphics[width=\linewidth]{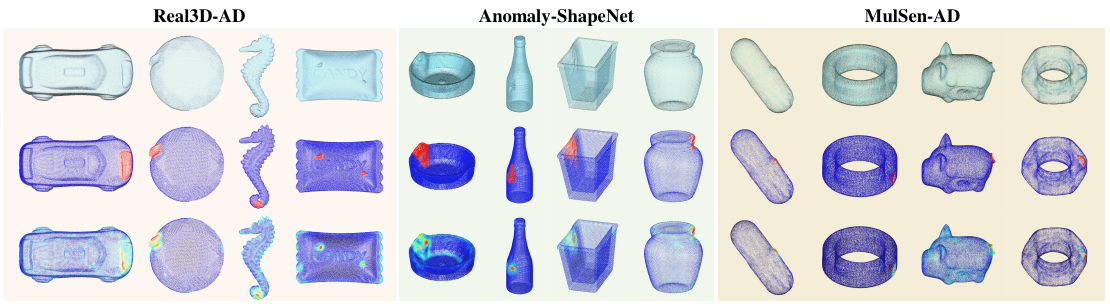}
\caption{\textbf{Visualization of our point-wise anomaly detection results on Real3D-AD, Anomaly-ShapeNet, and MulSen-AD.} From top to bottom: input point clouds, ground truths, anomaly maps.}
\vspace{-3mm}
\label{fig:vis_public}
\end{figure*}

\subsection{Benchmarking Results on MiniShift}

The quantitative assessment of all competing methods on MiniShift is delineated in Table~\ref{table:minishift}. Our approach consistently surpasses all baseline methods across diverse difficulty tiers, underscoring its exceptional generalization and robustness. As the dataset complexity escalates, existing methods manifest a marked deterioration in performance, particularly in the most arduous scenarios, where their efficacy in both O-ROC and P-ROC metrics approaches futility. Conversely, while our method also incurs a modest decline in performance, it sustains a decisive and substantial advantage over all rival techniques. This superior performance is chiefly ascribed to two pivotal components of our methodology, which enable the extraction of highly discriminative and expressive features. As a result, our approach adeptly discerns even the most subtle and intricate anomalies that elude detection by other methods.

The qualitative analysis, as depicted in Figure~\ref{fig:vis_minishift}, corroborates these quantitative outcomes. Our method, \ourmethod{}, produces anomaly maps of remarkable sharpness and precision across multiple categories, even amidst intricate and challenging conditions. In stark contrast, the majority of baseline methods struggle to delineate defects from the background effectively, thereby underscoring the critical necessity of high-resolution representations and meticulous local feature descriptions for proficient anomaly detection. Detailed per-category evaluations and additional qualitative illustrations are furnished in \appen{C} and \appen{H}, respectively.

\subsection{Benchmarking Results on Existing Datasets}

To further ascertain the efficacy of \ourmethod{}, we extend our evaluation to three prominent public benchmarks: Real3D-AD, Anomaly-ShapeNet, and MulSen-AD. The outcomes are encapsulated in Table~\ref{table:public_results}. \ourmethod{} uniformly sets new SOTA benchmarks across all datasets for both O-ROC and P-ROC metrics. Moreover, our method exhibits robust per-category performance and precise localization of fine-grained anomalies, thereby affirming its generalizability and discriminative prowess across a spectrum of 3D anomaly types.

\noindent\textbf{Real3D-AD:}
On the Real3D-AD dataset, \ourmethod{} attains scores of 80.4\% and 92.3\% for O-ROC and P-ROC, respectively, thereby eclipsing prior SOTA methods such as MC3D-AD and ISMP by margins of 2.2\% and 8.7\% in the respective metrics. The comprehensive per-category analysis, detailed in \appen{D}, reveals that \ourmethod{} secures the highest object-wise or point-wise performance in 10 out of 12 categories, thereby attesting to its efficacy. Particularly noteworthy is the near-perfect detection achieved in the Diamond (100\%/99.0\%) and Car (98.1\%/99.2\%) categories.

\noindent\textbf{Anomaly-ShapeNet:}
For the Anomaly-ShapeNet dataset, \ourmethod{} records O-ROC and P-ROC scores of 86.0\% and 92.9\%, respectively. Significantly, our method not only outperforms MC3D-AD by 1.8\% in O-ROC but also demonstrates a substantial enhancement of 18.1\% in P-ROC. The per-category breakdown, presented in \appen{E}, indicates that \ourmethod{} achieves superior performance in 35 out of 40 categories and attains complete detection in 5 categories, further solidifying its preeminence.

\begin{figure*}[t!]
\centering\includegraphics[width=\linewidth]{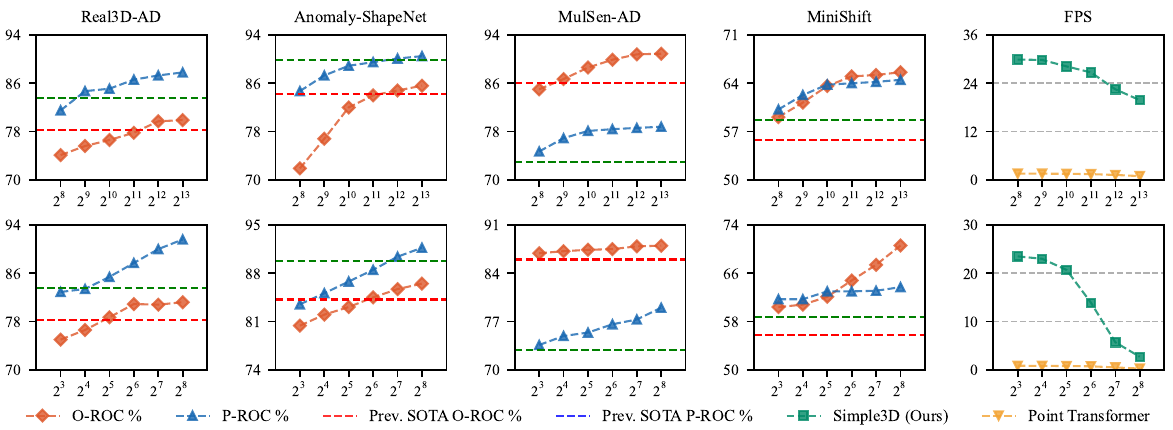}
\caption{\textbf{Influence of detection resolution $t$ (First Row) and aggregated neighborhood point number $k_L$ (Second Row).} The left four figures show the performance on datasets Real3D-AD, Anomaly-ShapeNet, MulSen-AD, and MiniShift, while the last one shows the Frames Per Second (FPS), as the sampling resolution $t$ varies from 256 ($2^8$) to 8192 ($2^{13}$) and aggregated neighborhood point number $k_L$ varies from 8 ($2^3$) to 128 ($2^8$). The FPS is statistically analyzed on the Anomaly-ShapeNet.}
\vspace{0mm}
\label{fig:ablation}
\end{figure*}

\noindent\textbf{MulSen-AD:}
On the MulSen-AD dataset, \ourmethod{} establishes a new SOTA benchmark with scores of 88.3\% and 80.3\% for O-ROC and P-ROC, respectively, thereby surpassing the previous best by 2.2\% and 16.2\%. The per-category evaluation, as documented in \appen{F}, reveals that our method attains the highest point-wise performance in 12 out of 15 categories.

\noindent\textbf{Qualitative Results and Analysis:}
The qualitative results, as showcased in Figure~\ref{fig:vis_public}, exemplify the adeptness of \ourmethod{} in precisely localizing subtle anomalies across the Real3D-AD, Anomaly-ShapeNet, and MulSen-AD datasets. These visualizations accentuate the method's superior detection capabilities, particularly when juxtaposed with the shortcomings exhibited by competing approaches.

\subsection{Ablation Study}
To elucidate the impact of MSND and LFSA, we undertake a thorough ablation study across all four datasets. The results, as tabulated in Table~\ref{table:Ablation}, demonstrate that both components invariably augment detection performance, yielding substantial gains across all evaluated metrics.

\begin{table}[]
\centering
\caption{\textbf{Ablation results of MSND and LFSA between four datasets.} }
\label{table:Ablation}
\fontsize{11}{14}\selectfont{
\resizebox{\linewidth}{!}{
\begin{tabular}{c|c|c|c|c}
\toprule[1.5pt]
MSND  & \ding{55} &  \ding{51} & \ding{55} &  \ding{51} \\
LFSA  &  \ding{55} &  \ding{55} & \ding{51} &  \ding{51}\\
 \midrule
Real3D-AD & 74.2/76.9 & 78.1/85.3 & 79.8/90.0 & 80.4/92.3 \\
Anomaly-ShapeNet & 78.8/77.0 & 83.8/85.6 & 85.7/90.4 & 86.0/92.9 \\
MulSen-AD & 85.1/66.2 & 86.8/70.7& 87.1/77.3 & 88.2/80.3 \\
MiniShift & 67.2/56.7 & 67.8/57.9 & 67.4/63.1 & 68.6/66.2 \\
 \bottomrule[1.5pt]

\end{tabular}}
}

\vspace{-5mm}
\end{table}
 
\subsubsection{Analysis on Detection Resolution}
To delve deeper into the influence of detection resolution on efficacy, we assess our method across a spectrum of resolutions ranging from $2^8$ to $2^{13}$, while maintaining a fixed aggregation point count of 16, utilizing the four benchmark datasets. The findings, depicted in the first row of Figure~\ref{fig:ablation}, reveal a consistent enhancement in all evaluation metrics concomitant with increased resolution across all datasets. Notably, at resolutions exceeding $2^{12}$, our method surpasses the performance of prior SOTA techniques. These observations underscore the pivotal importance of high-resolution representations in bolstering the accuracy and dependability of anomaly detection.

\subsubsection{Analysis on Number of Aggregated Points}
\ourmethod{} incorporates LFSA to expand the receptive field of local features for better feature distinguishability. To quantify the effect of the aggregated point count on detection performance, we systematically vary the neighborhood size from $2^3$ to $2^8$ points, while keeping the input resolution constant at 4k points. The outcomes, illustrated in the second row of Figure~\ref{fig:ablation}, demonstrate a uniform improvement in performance metrics across all datasets as the number of aggregated points increases. These findings accentuate the indispensable role of meticulous local feature description in 3D anomaly detection, with our approach occasionally outperforming features derived from learned 3D backbones.

\subsubsection{Analysis on Runtime}
The runtime analysis, as presented in Figure~\ref{fig:ablation}, elucidates the influence of resolution and the number of aggregation points on detection speed. Notably, at a resolution of 4k points, \ourmethod{} not only surpasses prior SOTA methods in performance but also sustains a detection speed exceeding 20 FPS, markedly outpacing all competing approaches. This real-time processing capability is of paramount significance for practical applications, such as industrial inspection.

\section{Conclusion}

We present \textbf{MiniShift}, a novel benchmark for high-resolution 3D anomaly detection, constructed via a controllable synthetic pipeline that scalably generates realistic and subtle surface defects. Our evaluation protocol classifies anomalies into three difficulty levels based on geometric saliency and perceptual visibility, ensuring a comprehensive assessment aligned with real-world industrial complexities. 
Current SOTA methods struggle to balance accuracy and efficiency in high-resolution contexts. To overcome this, we propose \textbf{Simple3D}, a lightweight yet potent method enabling real-time, fine-grained anomaly detection in dense point clouds. Rigorous experiments on \textbf{MiniShift} and datasets such as Real3D-AD, Anomaly-ShapeNet, and MulSenAD reveal that \textbf{Simple3D} achieves SOTA performance in both accuracy and speed, emphasizing the essential role of high-resolution representations in detecting minute defects. 
We assert that \textbf{MiniShift} and \textbf{Simple3D} lay robust groundwork for future advancements in high-resolution 3D anomaly detection.

\noindent \textbf{Future Directions}: Leveraging our scalable anomaly generation pipeline, we aim to construct a large-scale, diverse dataset for 3D anomaly detection, facilitating the development of foundational models adept at identifying subtle anomalies across varied categories for seamless industrial integration.



\bibliography{aaai2026}

\clearpage
\appendix

\section{Appendix}
The supplementary material includes the following sections to provide additional support for the main manuscript:

\begin{itemize}

    \item \textbf{Sec. \textcolor{red}{A}}: Details about MiniShift
    \item \textbf{Sec. \textcolor{red}{B}}: Dataset Comparison
    \item \textbf{Sec. \textcolor{red}{C}}: Detailed Results on MiniShift
    \item \textbf{Sec. \textcolor{red}{D}} Detailed Results on Real3D-AD
    \item \textbf{Sec. \textcolor{red}{E}} Detailed Results on Anoamly-ShapeNet
    \item \textbf{Sec. \textcolor{red}{F}} Detailed Results on MulSen-AD
    \item \textbf{Sec. \textcolor{red}{G}} More Visualization of MiniShift
    \item \textbf{Sec. \textcolor{red}{H}} More Qualitative Results
    
\end{itemize}

\begin{table*}[t]
\centering
\caption{\textbf{Detailed data distribution for each category in \ourdataset{}.}}
\label{tab:statistic}
\resizebox{0.8\linewidth}{!}{
\begin{tabular}{l|c|ccccc|c|c} 
\toprule[1.5pt]
\multirow{2}{*}{\textbf{Category}} &
  \multicolumn{1}{c|}{\#\textbf{Training}} &
  \multicolumn{5}{c|}{\#\textbf{Testing}} &
  \multirow{2}{*}{\textbf{Total}} &

  \multirow{2}{*}{\textbf{Resolution}}  \\ \cmidrule(lr){2-7} 
 &
  \textit{Normal} &
  \textit{Normal} &
  \textit{Areal} &
  \textit{Striate} &
  \textit{Scratch} &
  \textit{Sphere} & \\ \midrule
Capsule &  54 &  10 &  30 &  30 &  30 &  30 &  184 &  500,000 \\ 
Cube &  100 &  10 &  30 &  30 &  30 &  30 &  230 &  500,000 \\ 
Spring pad &  76 &  10 &  30 &  30 &  30 &  30 &  206 &  500,000 \\ 
Screw &  80 &  10 &  30 &  30 &  30 &  30 &  210 &  500,000 \\ 
Screen &  59 &  10 &  30 &  30 &  30 &  30 &  189 &  500,000 \\ 
Piggy &  100 &  10 &  30 &  30 &  30 &  30 &  230 &  500,000 \\ 
Nut &  108 &  10 &  30 &  30 &  30 &  30 &  238 &  500,000 \\ 
Flat pad &  80 &  10 &  30 &  30 &  30 &  30 &  210 &  500,000 \\ 
Plastic   cylinder &  80 &  10 &  30 &  30 &  30 &  30 &  210 &  500,000 \\ 
Button   cell &  80 &  10 &  30 &  30 &  30 &  30 &  210 &  500,000 \\ 
Toothbrush &  100 &  10 &  30 &  30 &  30 &  30 &  230 &  500,000 \\ 
Light &  100 &  10 &  30 &  30 &  30 &  30 &  230 &  500,000 \\

\bottomrule[1.5pt] 
\end{tabular}
}
\end{table*}

\subsection{\textcolor{red}{A}. Details about MiniShift}~\label{Details_MiniShift}

Table~\ref{tab:statistic} presents the detailed data distribution for each category in \ourdataset{}. \ourdataset{} comprises 12 categories (Capsule, Cube, Spring pad, Screw, Screen, Piggy, Nut, Flat pad, Plastic cylinder, Button cell, Toothbrush, and Light.), with a total of 2577 samples. Specifically, we selected 10 3D samples from the corresponding training data of MulSenAD for synthetic anomaly generation, while the remaining samples are used as the our training set. These 10 samples are utilized to generate 120 anomalous instances to form the testing set, with 30 samples for each defect type. The testing set also includes 10 normal samples from the MulSen testing set, resulting in a total of 130 test samples for each category.

\subsection{\textcolor{red}{B}. Dataset Comparison}~\label{Dataset_Comparison}

Statistical comparisons between \ourdataset{} and existing 3D AD datasets, are summarized in Table~\ref{tab:dataset_comparison}, highlighting that our dataset offers the highest spatial resolution (500,000 points significantly higher than other datasets) and exhibits the lowest anomaly coverage ($<1.0\%$ significantly more challenging than other datasets).

\begin{table*}[t]
\centering
\caption{\textbf{Statistical comparisons between \ourdataset{} and existing 3D AD datasets.} \ourdataset{} dataset offers the highest spatial resolution and exhibits the lowest anomaly coverage, posing greater challenges for fine-grained anomaly detection.}
\label{tab:dataset_comparison}
\resizebox{0.8\linewidth}{!}{
\begin{tabular}{l|c|ccc|c|c} 
\toprule[1.5pt]
\multirow{2}{*}{\textbf{Dataset}} &
  \multirow{2}{*}{\#\textbf{Category}} &
  \multicolumn{3}{c|}{\#\textbf{Point Cloud}} &
  \multirow{2}{*}{\textbf{Resolution}} &
  \multirow{2}{*}{\textbf{Ratio}}
   \\ \cmidrule(lr){3-5} 
 & &
  \textit{Normal} &
  \textit{Abnormal} &
  \textit{Total} &
 \\ \midrule
MVTec 3D-AD &  10 &  2,904 & 948 &  3,852 &  26,532 &  1.8\%  \\ 
Real3D-AD &  12 &  652 &  602 &  1,254 &  169,022 &  2.7\%  \\ 
Anomaly-ShapeNet &  40 &  160 &  1,440 &  1,600 &  58,604 &  2.4\%  \\ 
MulSen-AD &  15 &  1,541 &  494 &  2,035 &  29,818 &  4.2\% \\ 
MiniShift & 12 & 1137 & 1,200 & 2577 & 500,000 & 0.99\%  \\ 
  
\bottomrule[1.5pt] 
\end{tabular}
}
\end{table*}

\subsection{\textcolor{red}{C}. Detailed Results on MiniShift}~\label{Results_on_MiniShift}

Tabel~\ref{table:minishift_all} shows the detailed performance of each category. \ourmethod{} achieves the best performance in 10 of 12 categories, and is significantly superior to other methods in the average performance. The results of the three subsets also consistently support the superiority of our method, as shown in Table Tabel~\ref{table:minishift_easy}, Tabel~\ref{table:minishift_medium}, and Tabel~\ref{table:minishift_hard}.
Nevertheless, the O-ROC/P-ROC less than 70\% highlights the challenge of detecting subtle defects. 

\begin{table}[t]
\centering
\caption{\textbf{Quantitative Results on MiniShift-ALL}. The results are presented in O-ROC\%/P-ROC\%. The best performance is in \textbf{bold}, and the second best is \underline{underlined}.}
\label{table:minishift_all}
\fontsize{11}{14}\selectfont{
\resizebox{\linewidth}{!}{
\begin{tabular}{c|ccccc|
>{\columncolor{blue!8}}c}
\toprule[1.5pt]

Method~$\rightarrow$  &  \small{PatchCore-FP} &  \small{PatchCore-PM} &  \small{PatchCore-PB}  & \small{R3D-AD}  &  \small{GLFM}   &  \textbf{\ourmethod}  \\
Category~$\downarrow$   & \small{CVPR'22} &  \small{CVPR'22} & \small{CVPR'22} & \small{ECCV'24} &  \small{TASE'25} &  \textbf{Ours}   \\ \midrule

Capsule            & \underline{88.0}/58.4 & 32.4/54.6 & 27.2/51.1 & 60.0/51.3 & 41.2/\underline{69.4} & \textbf{92.9}/\textbf{80.6} \\
Cube               & \textbf{61.9}/52.7 & 37.5/51.2 & 44.1/47.7 & 53.5/53.7 & 48.8/\underline{65.5} & \textbf{61.9}/\textbf{66.6} \\
Spring pad         & \textbf{68.9}/53.4 & 53.2/\underline{55.0} & 41.0/52.8 & 32.9/50.3 & 53.4/54.5 & \underline{68.7}/\textbf{63.1} \\
Screw              & 25.0/\underline{56.0} & 59.2/54.9 & \underline{67.2}/51.3 & 63.9/48.7 & \textbf{68.2}/53.4 & 23.2/\textbf{64.6} \\
Screen             & \textbf{70.2}/53.3 & 63.5/52.8 & 54.0/50.2 & 42.9/47.2 & 63.9/\textbf{66.8} & \underline{70.0}/\underline{57.5} \\
Piggy              & 61.3/\underline{55.5} & 58.4/53.8 & 50.3/52.4 & 36.0/49.7 & \underline{63.8}/53.4 & \textbf{67.4}/\textbf{67.5} \\
Nut                & \underline{69.2}/54.5 & 58.0/55.1 & 43.2/53.0 & 66.3/53.7 & 49.4/\underline{56.6} & \textbf{76.9}/\textbf{73.1} \\
Flat pad           & 60.7/54.1 & 59.9/\underline{57.8} & 60.9/52.7 & 59.3/49.5 & \underline{67.7}/57.6 & \textbf{68.8}/\textbf{61.8} \\
Plastic   cylinder & \underline{81.2}/54.8 & 77.7/53.0 & 60.3/51.6 & 70.2/50.7 & 63.5/\underline{58.9} & \textbf{83.2}/\textbf{70.3} \\
Button   cell      & \underline{86.2}/57.4 & 64.3/52.7 & 51.1/48.4 & 43.7/50.3 & 72.0/\underline{62.3} & \textbf{90.0}/\textbf{73.2} \\
Toothbrush         & 48.8/46.7 & 55.1/\textbf{56.7} & 49.0/\underline{55.2} & \textbf{66.3}/39.2 & 44.1/47.8 & \underline{66.1}/45.5 \\
Light              & \textbf{59.8}/47.7 & 51.2/51.6 & 51.3/50.5 & 48.2/53.0 & 33.0/\underline{58.2} & \underline{54.2}/\textbf{70.1} \\ \hline
Mean               & \underline{65.1}/53.7 & 55.9/54.1 & 50.0/51.4 & 53.6/49.8 & 55.8/\underline{58.7} & \textbf{68.6}/\textbf{66.2}
\\

\bottomrule[1.5pt]

\end{tabular}}}
\end{table}
\begin{table}[h!]
\centering
\caption{\textbf{Quantitative Results on MiniShift-easy}. The results are presented in O-ROC\%/P-ROC\%. The best performance is in \textbf{bold}, and the second best is \underline{underlined}.}
\label{table:minishift_easy}
\fontsize{11}{14}\selectfont{
\resizebox{\linewidth}{!}{
\begin{tabular}{c|ccccc|
>{\columncolor{blue!8}}c}
\toprule[1.5pt]

Method~$\rightarrow$  &  \small{PatchCore-FP} &  \small{PatchCore-PM} &  \small{PatchCore-PB}  & \small{R3D-AD}  &  \small{GLFM}   &  \textbf{\ourmethod}  \\
Category~$\downarrow$   & \small{CVPR'22} &  \small{CVPR'22} & \small{CVPR'22} & \small{ECCV'24} &  \small{TASE'25} &  \textbf{Ours}    \\ \midrule

Capsule            & \underline{99.8}/66.9 & 46.0/56.7 & 43.2/52.9 & 50.5/49.3 & 61.2/\underline{87.4} & \textbf{100.0}/\textbf{97.1} \\
Cube               & 62.8/54.2 & 44.8/52.9 & \underline{64.8}/46.8 & 41.8/49.3 & 47.0/\underline{74.1} & \textbf{72.5}/\textbf{78.2}  \\
Spring pad         & \underline{71.0}/54.7 & 43.8/\underline{62.3} & 46.5/52.3 & 53.5/51.6 & 41.5/59.1 & \textbf{72.2}/\textbf{70.2}  \\
Screw              & 25.0/53.8 & \underline{65.5}/57.2 & 56.2/55.4 & 61.8/52.6 & \textbf{69.0}/\underline{58.3} & 25.2/\textbf{73.2}  \\
Screen             & 69.5/55.2 & 67.2/54.4 & 54.8/47.6 & 61.0/43.1 & \textbf{76.7}/\textbf{81.4} & \underline{73.0}/\underline{67.1}  \\
Piggy              & 62.0/56.8 & \textbf{70.2}/56.9 & 62.5/55.3 & \underline{69.8}/46.3 & 54.8/\underline{62.3} & \underline{69.8}/\textbf{78.4}  \\
Nut                & \underline{74.0}/56.4 & 60.7/57.1 & 49.2/52.1 & 60.8/47.4 & 45.0/\underline{61.7} & \textbf{85.8}/\textbf{87.0}  \\
Flat pad           & 62.5/55.6 & \underline{64.5}/62.9 & 51.0/51.1 & 59.5/50.1 & 63.8/\underline{63.9} & \textbf{77.8}/\textbf{68.2}  \\
Plastic   cylinder & \underline{84.5}/55.8 & 72.2/55.6 & 69.5/49.8 & 45.8/49.9 & 70.0/\underline{68.4} & \textbf{91.5}/\textbf{85.1}  \\
Button   cell      & \underline{95.0}/67.4 & 62.8/57.7 & 50.5/49.1 & 50.0/50.2 & 65.0/\underline{74.2} & \textbf{95.7}/\textbf{93.2}  \\
Toothbrush         & 51.8/48.6 & 48.0/\textbf{58.3} & \underline{69.5}/53.9 & 52.5/33.8 & 61.8/\underline{54.9} & \textbf{75.2}/50.0  \\
Light              & 62.0/52.6 & 41.2/55.8 & 48.0/52.4 & \underline{68.5}/55.8 & 37.0/\underline{67.1} & \textbf{69.0}/\textbf{80.3}  \\ \hline
Mean               & \underline{68.3}/56.5 & 57.2/57.3 & 55.5/51.6 & 56.3/48.3 & 57.7/\underline{67.7} & \textbf{75.6}/\textbf{77.3}  \\

\bottomrule[1.5pt]

\end{tabular}}}
\end{table}

\begin{table}[h!]
\centering
\caption{\textbf{Quantitative Results on MiniShift-Medium}. The results are presented in O-ROC\%/P-ROC\%. The best performance is in \textbf{bold}, and the second best is \underline{underlined}.}
\label{table:minishift_medium}
\fontsize{11}{14}\selectfont{
\resizebox{\linewidth}{!}{
\begin{tabular}{c|ccccc|
>{\columncolor{blue!8}}c}
\toprule[1.5pt]

Method~$\rightarrow$  &  \small{PatchCore-FP} &  \small{PatchCore-PM} &  \small{PatchCore-PB}  & \small{R3D-AD}  &  \small{GLFM}   &  \textbf{\ourmethod}  \\
Category~$\downarrow$   & \small{CVPR'22} &  \small{CVPR'22} & \small{CVPR'22} & \small{ECCV'24} &  \small{TASE'25} &  \textbf{Ours}   \\ \midrule

Capsule            & \underline{88.8}/55.2 & 48.8/54.5 & 35.0/49.3 & 50.0/54.6 & 38.2/\underline{59.7} & \textbf{97.2}/\textbf{84.2} \\
Cube               & \textbf{63.2}/55.7 & 36.0/51.1 & 34.2/51.8 & 58.3/55.9 & 49.5/\underline{65.0} & \underline{59.0}/\textbf{69.2} \\
Spring pad         & \underline{68.0}/\underline{53.3} & 47.0/51.7 & 43.7/50.5 & 46.8/46.6 & 54.2/49.7 & \textbf{69.3}/\textbf{64.6} \\
Screw              & 25.0/\textbf{60.3} & 62.0/48.5 & \underline{67.5}/45.9 & 53.8/48.8 & \textbf{72.2}/50.1 & 25.8/\underline{56.9} \\
Screen             & \textbf{71.2}/53.8 & 69.5/48.6 & 54.8/49.3 & 59.0/50.6 & \textbf{71.2}/\textbf{67.3} & 66.5/\underline{55.1} \\
Piggy              & \underline{60.0}/\underline{59.0} & 52.3/53.2 & 44.5/46.7 & 54.8/51.5 & 56.2/55.1 & \textbf{72.2}/\textbf{67.6} \\
Nut                & \underline{71.5}/52.7 & 58.8/52.2 & 49.5/51.4 & 56.0/47   & 42.8/\underline{54.1} & \textbf{79.0}/\textbf{70.3} \\
Flat pad           & 59.8/54.6 & \underline{63.5}/57.0 & 40.2/52.6 & 54.3/52.8 & 58.2/\underline{58.7} & \textbf{65.2}/\textbf{60.9} \\
Plastic   cylinder & \underline{83.5}/57.3 & 79.0/58.6 & 42.5/51.2 & 34.5/51.1 & 65.5/\underline{61.5} & \textbf{86.0}/\textbf{69.4} \\
Button   cell      & \underline{85.5}/54.6 & 65.8/49.2 & 50.8/47.8 & 49.8/51.6 & 58.8/\underline{62.3} & \textbf{90.5}/73.3 \\
Toothbrush         & 52.2/46.1 & 48.8/\underline{54.1} & \underline{54.8}/\textbf{54.2} & 33.0/45.4 & 52.8/45.8 & \textbf{59.7}/45.1 \\
Light              & \underline{59.0}/46.4 & 37.7/52.6 & 42.0/53.4 & \textbf{59.3}/52   & 40.5/\underline{58.1} & 53.0/\textbf{69.9} \\ \hline
Mean               & \underline{65.6}/54.1 & 55.8/52.6 & 46.6/50.3 & 50.8/50.7 & 55.0/\underline{57.3} & \textbf{68.6}/\textbf{65.5}
\\


\bottomrule[1.5pt]

\end{tabular}}}
\end{table}
\begin{table}[t]
\centering
\caption{\textbf{Quantitative Results on MiniShift-Hard}. The results are presented in O-ROC\%/P-ROC\%. The best performance is in \textbf{bold}, and the second best is \underline{underlined}.}
\label{table:minishift_hard}
\fontsize{11}{14}\selectfont{
\resizebox{\linewidth}{!}{
\begin{tabular}{c|ccccc|
>{\columncolor{blue!8}}c}
\toprule[1.5pt]

Method~$\rightarrow$  &  \small{PatchCore-FP} &  \small{PatchCore-PM} &  \small{PatchCore-PB}  & \small{R3D-AD}  &  \small{GLFM}   &  \textbf{\ourmethod}  \\
Category~$\downarrow$   & \small{CVPR'22} &  \small{CVPR'22} & \small{CVPR'22} & \small{ECCV'24} &  \small{TASE'25} &   \textbf{Ours}    \\ \midrule

Capsule            & \underline{75.5}/54.4 & 44.7/50.0 & 40.2/49.4 & 56.0/50.6 & 39.2/\underline{61.6} & \textbf{81.5}/\textbf{62.8} \\
Cube               & \textbf{59.8}/52.4 & 34.2/52.6 & 40.2/51.9 & 55.8/\textbf{57.4} & 37.2/54.2 & \underline{54.3}/\underline{55.7} \\
Spring pad         & \textbf{67.8}/52.1 & 48.5/53.0 & 36.2/49.6 & 38.3/48.4 & 63.2/\underline{54.8} & \underline{64.5}/\textbf{55.5} \\
Screw              & 25.0/52.6 & 56.8/54.2 & \textbf{63.5}/52.9 & 34.5/48.1 & \underline{59.0}/\underline{54.8} & 18.8/\textbf{57.9} \\
Screen             & \underline{69.8}/51.0 & 64.5/56.3 & 62.8/\underline{58.1} & 39.3/53.2 & 60.8/\textbf{59.3} & \textbf{70.5}/51.5 \\
Piggy              & 62.0/\underline{50.1} & \underline{62.2}/48.5 & \textbf{65.8}/46.7 & 56.3/45.7 & 58.0/39.0 & 60.2/\textbf{52.9} \\
Nut                & \underline{62.0}/54.7 & 61.8/48.5 & 51.5/51.2 & 60.8/53.9 & 53.5/\underline{59.3} & \textbf{66.0}/\textbf{62.4} \\
Flat pad           & 60.0/53.7 & \textbf{63.8}/\textbf{59.4} & 54.0/50.7 & 45.8/50.7 & 62.5/51.3 & \underline{63.3}/\underline{59.2} \\
Plastic   cylinder & \textbf{75.5}/\underline{51.1} & 67.0/50.9 & 59.2/45.5 & 51.0/47.0 & 62.8/49.1 & \underline{72.2}/\textbf{58.0} \\
Button   cell      & \underline{78.0}/50.8 & 65.0/48.2 & 62.8/46.7 & 31.8/\underline{54.6} & 59.8/53.0 & \textbf{83.8}/\textbf{56.5} \\
Toothbrush         & 42.5/45.9 & 45.5/\textbf{59.5} & 48.2/\underline{58.0} & \underline{61.5}/37.6 & 45.8/44.1 & \textbf{63.2}/41.7 \\
Light              & \textbf{58.5}/44.3 & 38.0/42.9 & 49.8/49.4 & \underline{56.8}/\underline{54.3} & 32.2/52.1 & 40.8/\textbf{61.4} \\ \hline
Mean               & \underline{61.4}/51.1 & 54.3/52.0 & 52.9/50.8 & 50.1/50.6 & 52.8/\underline{52.7} & \textbf{61.6}/\textbf{56.3}
\\


\bottomrule[1.5pt]

\end{tabular}}}
\end{table}

\subsection{\textcolor{red}{D}. Detailed Results on Real3D-AD}~\label{Results_on_Real3D}
Table 11 presents the quantitative comparison on the Real3D-AD dataset. Simple3D, achieves the best performance across the board, with an average O-ROC of 80.4\% and P-ROC of 92.3\%, outperforming all existing baselines.  Notably, Simple3D ranks first in 7 out of 12 categories for P-ROC, including Car (99.2\%), Diamond (99.0\%), Seahorse (94.2\%), and Fish (96.2\%), while maintaining competitive or top-tier results in most other classes.

\begin{table*}[t]
\centering
\caption{\textbf{Quantitative Results on Real3D-AD.} The results are presented in O-ROC\%/P-ROC\%. The best performance is in \textbf{bold}, and the second best is \underline{underlined}.}
\label{table:real}
\fontsize{11}{14}\selectfont{
\resizebox{\linewidth}{!}{
\begin{tabular}{c|cccccccc|
>{\columncolor{blue!8}}c}
\toprule[1.5pt]

Method~$\rightarrow$   & \small{CPMF} & 
\small{Reg3D-AD} & \small{Group3AD}   & \small{IMRNet}   & \small{ISMP}  & \small{GLFM}    &  \small{PO3AD}  &  \small{MC3D-AD}   & \textbf{Simple3D}    \\
Real3D-AD  & \small{PR'24} & \small{NeurIPS'23} & \small{ACM MM'24}   & \small{CVPR'24} & \small{AAAI'25} & \small{TASE'25} & \small{CVPR'25} & \small{IJCAI'25} & \textbf{Ours}   \\ \midrule

Airplane & 63.2/61.8 & 71.6/63.1 & 74.4/63.6 & 76.2/- & \textbf{85.8}/\underline{75.3} & 54.6/74.3 & 80.4/- & \underline{85.0}/62.8 & 76.5/\textbf{88.1} \\
Car      & 51.8/83.6 & 69.7/71.8 & 72.8/74.5 & 71.1/- & 73.1/83.6 & \underline{84.2}/\underline{88.2} & 65.4/- & 74.9/81.9 & \textbf{98.1}/\textbf{99.2} \\
Candybar & 71.8/73.4 & 82.7/72.4 & \underline{84.7}/73.8 & 75.5/- & \textbf{85.2}/90.7 & 71.5/79.7 & 78.5/- & 83.0/\underline{91.0} & 65.1/\textbf{96.2} \\
Chicken  & 64.0/55.9 & \textbf{85.2}/67.6 & 78.6/75.9 & 78.0/- & 71.4/\underline{79.8} & 68.8/62.4 & 68.6/- & 71.5/64.0 & \underline{82.6}/\textbf{86.1} \\
Diamond  & 64.0/75.3 & 90.0/83.5 & 93.2/86.2 & 90.5/- & 94.8/92.6 & 71.2/76.8 & 80.1/- & \underline{95.5}/\underline{94.2} & \textbf{100}/\textbf{99.0}  \\
Duck     & 55.4/71.9 & 58.4/50.3 & 67.9/63.1 & 51.7/- & 71.2/\underline{87.6} & \textbf{94.5}/66.3 & 82.0/- & \underline{83.1}/82.2 & 77.8/\textbf{96.6} \\
Fish     & 84.0/\textbf{98.8} & 91.5/82.6 & \textbf{97.6}/83.6 & 88.0/- & \underline{94.5}/88.6 & 69.5/94.2 & 85.9/- & 86.5/93.2 & 91.2/\underline{96.2} \\
Gemstone & 34.9/44.9 & 41.7/54.5 & 53.9/56.4 & 67.4/- & 46.8/\underline{85.7} & 68.8/75.0 & \underline{69.3}/- & 56.0/45.8 & \textbf{70.4}/\textbf{97.3} \\
Seahorse & 84.3/\textbf{96.2} & 76.2/81.7 & 84.1/82.7 & 60.4/- & 72.9/81.3 & \underline{92.4}/81.5 & 75.6/- & 71.6/65.9 & \textbf{93.0}/\underline{94.2} \\
Shell    & 39.3/72.5 & 58.3/\underline{81.1} & 58.5/79.8 & 66.5/- & 62.3/\textbf{83.9} & 73.3/60.2 & \underline{80.0}/- & \textbf{80.3}/77.8 & 51.4/71.6 \\
Starfish & 52.6/\underline{80.0} & 50.6/61.7 & 56.2/62.5 & 67.4/- & 66.0/64.1 & 74.8/67.5 & \underline{75.8}/- & \textbf{76.6}/69.0 & 69.6/\textbf{85.8} \\
Toffees  & \underline{84.5}/\underline{95.9} & 68.5/75.9 & 79.6/80.3 & 77.4/- & 84.2/89.5 & 76.3/93.5 & 77.1/- & 73.8/93.4 & \textbf{88.8}/\textbf{96.8} \\ \hline
Mean     & 62.5/75.9 & 70.4/70.5 & 75.1/73.5 & 72.5/- & 76.7/\underline{83.6} & 75.0/76.7 & 76.5/- & \underline{78.2}/76.8 & \textbf{80.4}/\textbf{92.3} \\ \bottomrule[1.5pt]
\end{tabular}}}
\end{table*}

\subsection{\textcolor{red}{E}. Detailed Results on Anoamly-ShapeNet}~\label{Results_on_ShapeNet}

Quantitative comparison on the Anoamly-ShapeNet dataset is provided in Tabel~\ref{table:shapenet}. Compared to recent SOTA methods such as MC3D-AD and PO3AD, Simple3D exhibits superior object- and point-level detection and robustness across diverse object categories. For example, although MC3D-AD performs well in O-ROC for some categories, its average P-ROC scores often lag behind, highlighting weaker spatial precision. In contrast, Simple3D delivers consistently high P-ROC values, validating its strength in dense and fine-grained anomaly perception.

\begin{table*}[t]
\centering
\caption{\textbf{Quantitative Results on Anomaly-ShapeNet}. The results are presented in O-ROC\%/P-ROC\%. The best performance is in \textbf{bold}, and the second best is \underline{underlined}.}
\label{table:shapenet}
\fontsize{11}{14}\selectfont{
\resizebox{\linewidth}{!}{
\begin{tabular}{c|cccccccc|
>{\columncolor{blue!8}}c
>{\columncolor{blue!8}}c
>{\columncolor{blue!8}}c
>{\columncolor{blue!8}}c
>{\columncolor{blue!8}}c
>{\columncolor{blue!8}}c}
\toprule[1.5pt]

Method~$\rightarrow$  & \small{M3DM-PM}  & \small{CPMF} & 
\small{Reg3D-AD}  & \small{IMRNet}   & \small{ISMP}   &  \small{GLFM}  &  \small{PO3AD}  &  \small{MC3D-AD}   & \textbf{Simple3D}     \\
Anomaly-ShapeNet  & \small{CVPR'23}  & \small{PR'24} & \small{NeurIPS'23} & \small{CVPR'24}   & \small{AAAI'25}  &  \small{TASE'25} & \small{CVPR'25} & \small{IJCAI'25} & \textbf{Ours}   \\ \midrule

Ashtray0    & 51.9/66.7 & 35.3/- & 59.7/- & 67.1/-    & -/60.3    & 52.8/74.1 & \textbf{100.0}/\textbf{96.2} & 96.2/-    & \underline{99.5}/\underline{92.0}   \\
Bag0        & 55.7/62.8 & 64.3/- & 70.6/- & 66.0/-    & -/74.7    & 53.7/75.4 & \underline{83.3}/\underline{94.9}  & 80.5/-    & \textbf{88.1}/\textbf{95.4} \\
Bottle0     & 49.5/54.5 & 52.0/- & 48.6/- & 55.2/-    & -/77.0    & 49.5/80.5 & \underline{90.0}/\underline{91.2}  & 79.5/-    & \textbf{97.6}/\textbf{97.4} \\
Bottle1     & 55.1/57.8 & 48.2/- & 69.5/- & 70.0/-    & -/56.8    & 69.9/70.7 & \underline{93.3}/\textbf{84.4}  & 70.9/-    & \textbf{95.1}/\underline{72.8} \\
Bottle3     & 44.1/69.5 & 40.5/- & 52.5/- & 64.0/-    & -/77.5    & 74.5/\underline{85.3} & \underline{92.6}/\textbf{88.0}  & 75.6/-    & \textbf{100}/83.8  \\
Bowl0       & 55.2/62.7 & 78.3/- & 67.1/- & 68.1/-    & -/85.1    & 53.1/83.0 & 92.2/\underline{97.8}  & \underline{93.0}/-    & \textbf{100}/\textbf{98.8}  \\
Bowl1       & 41.1/38.4 & 63.9/- & 52.5/- & 70.2/-    & -/54.6    & 54.3/62.0 & 82.9/\underline{91.4}  & \textbf{97.8}/-    & \underline{83.0}/\textbf{95.1}   \\
Bowl2       & 45.6/45.8 & 62.5/- & 49.0/- & 68.5/-    & -/73.6    & 61.0/73.8 & \textbf{83.3}/\underline{91.8}  & \underline{71.9}/-    & 71.1/\textbf{93.3} \\
Bowl3       & 40.7/45.9 & 65.8/- & 34.8/- & 59.9/-    & -/77.3    & 79.4/87.7 & 88.1/\underline{93.5}  & \underline{88.5}/-    & \textbf{91.1}/\textbf{99.3} \\
Bowl4       & 47.8/55.9 & 68.3/- & 66.3/- & 67.6/-    & -/74.0    & 75.1/59.5 & \textbf{98.1}/\textbf{96.7}  & \underline{91.1}/-    & 73.0/\underline{92.9}   \\
Bowl5       & 61.4/45.7 & 68.5/- & 59.3/- & 71.0/-    & -/53.4    & 72.0/58.2 & \underline{84.9}/\underline{94.1}  & 75.4/-    & \textbf{86.3}/\textbf{97.9} \\
Bucket0     & 53.3/42.1 & 48.2/- & 61.0/- & 58.0/-    & -/52.4    & 51.2/60.9 & 85.3/\textbf{75.5}  & \underline{89.8}/-    & \textbf{95.9}/\underline{72.5} \\
Bucket1     & 43.2/65.9 & 60.1/- & 75.2/- & 77.1/-    & -/67.2    & 66.2/69.4 & \underline{78.7}/\underline{89.9}  & 78.4/-    & \textbf{79.0}/\textbf{92.1}   \\
Cap0        & 48.5/68.1 & 60.1/- & 69.3/- & 73.7/-    & -/86.5    & 62.1/97.6 & \textbf{87.7}/\underline{95.7}  & 79.3/-    & \underline{85.2}/\textbf{98.8} \\
Cap3        & 58.6/65.5 & 55.1/- & 72.5/- & 77.5/-    & -/73.4    & 56.4/89.1 & \underline{85.9}/\underline{94.8}  & 70.1/-    & \textbf{86.7}/\textbf{96.4} \\
Cap4        & 61.1/54.2 & 55.3/- & 64.3/- & 65.2/-    & -/75.3    & 81.2/92.6 & 79.2/\underline{94.0}  & \underline{83.5}/-    & \textbf{91.2}/\textbf{97.9} \\
Cap5        & 49.5/50.7 & 69.7/- & 46.7/- & 65.2/-    & -/67.8    & 64.2/\underline{87.2} & 67.0/86.4  & \underline{76.1}/-    & \textbf{80.4}/\textbf{96.4} \\
Cup0        & 51.0/59.8 & 49.7/- & 51.0/- & 64.3/-    & -/86.9    & 56.4/73.1 & \underline{87.1}/\underline{90.9}  & 74.3/-    & \textbf{100}/\textbf{97.9}  \\
Cup1        & 42.4/51.5 & 49.9/- & 53.8/- & 75.7/-    & -/60.0    & 63.4/55.9 & \underline{83.3}/\underline{93.2}  & \textbf{95.2}/-    & 82.4/\textbf{93.7} \\
Eraser0     & 50.0/50.0 & 68.9/- & 34.3/- & 54.8/-    & -/70.6    & 55.7/60.9 & \underline{99.5}/\textbf{97.4}  & 77.6/-    & \textbf{100}/\underline{97.0}    \\
Headset0    & 49.8/50.4 & 64.3/- & 53.7/- & 72.0/-    & -/58.0    & 59.2/72.6 & 80.8/\underline{82.3}  & \underline{86.2}/-    & \textbf{98.2}/\textbf{93.6} \\
Headset1    & 50.0/55.7 & 45.8/- & 61.0/- & 67.6/-    & -/70.2    & 63.4/60.8 & \underline{92.3}/\underline{90.7}  & 88.6/-    & \textbf{95.7}/\textbf{95.8} \\
Helmet0     & 41.7/53.1 & 55.5/- & 60.0/- & 59.7/-    & -/68.3    & 59.3/75.3 & \textbf{76.2}/\underline{87.8}  & 67.2/-    & \underline{69.0}/\textbf{90.2}   \\
Helmet1     & 50.0/47.3 & 58.9/- & 38.1/- & 60.0/-    & -/62.2    & 62.8/62.6 & \underline{96.1}/\textbf{94.8}  & \textbf{100}/-     & 71.9/\underline{90.2} \\
Helmet2     & 55.1/66.4 & 46.2/- & 61.4/- & 64.1/-    & -/84.4    & 56.5/81.8 & \textbf{86.9}/\underline{93.2}  & 60.9/-    & \underline{75.4}/\textbf{94.7} \\
Helmet3     & 50.9/57.3 & 52.0/- & 36.7/- & 57.3/-    & -/72.2    & 53.8/72.8 & \underline{75.4}/\underline{84.6}  & \textbf{97.9}/-    & 65.5/\textbf{92.8} \\
Jar0        & 44.3/69.5 & 61.0/- & 59.2/- & 78.0/-    & -/82.3    & 56.9/74.9 & 86.6/\underline{87.1}  & \textbf{97.1}/-    & \underline{90.5}/\textbf{97.8} \\
Microphone0 & 53.3/59.7 & 50.9/- & 41.4/- & 75.5/-    & -/66.1    & 71.9/\underline{81.4} & 77.6/81.0  & \underline{91.9}/-    & \textbf{100}/\textbf{96.4}  \\
Shelf0      & 46.7/66.5 & 68.5/- & 68.8/- & 60.3/-    & -/\underline{68.7}    & 57.0/64.4 & 57.3/66.3  & \textbf{84.1}/-    & \underline{74.8}/\textbf{90.1} \\
Tap0        & 47.6/55.2 & 35.9/- & 67.6/- & 67.6/-    & -/52.2    & 68.9/70.6 & \underline{74.5}/\underline{78.3}  & \textbf{94.5}/-    & 70.3/\textbf{85.7} \\
Tap1        & 49.3/45.0 & 69.7/- & 64.1/- & 69.6/-    & -/55.2    & 42.6/\underline{78.3} & \underline{68.1}/69.2  & \textbf{97.0}/-    & 59.6/\textbf{81.1} \\
Vase0       & 52.9/50.7 & 45.1/- & 53.3/- & 53.3/-    & -/66.1    & 67.6/77.8 & \underline{85.8}/\textbf{95.5}  & 82.1/-    & \textbf{95.4}/\underline{93.4} \\
Vase1       & 66.2/71.9 & 34.5/- & 70.2/- & 75.7/-    & -/84.3    & 76.1/\textbf{93.4} & 74.2/\underline{88.2}  & \textbf{85.7}/-    & \underline{82.4}/80.7 \\
Vase2       & 70.0/55.1 & 58.2/- & 60.5/- & 61.4/-    & -/73.3    & 59.3/61.4 & \textbf{95.2}/\underline{97.8}  & \underline{92.9}/-    & 87.1/\textbf{98.6} \\
Vase3       & 50.9/51.5 & 58.2/- & 65.0/- & 70.0/-    & -/76.2    & 67.3/76.7 & \textbf{82.1}/\underline{88.4}  & 76.1/-    & \underline{81.2}/\textbf{91.0}   \\
Vase4       & 51.5/44.9 & 51.4/- & 50.0/- & 52.4/-    & -/54.5    & 57.3/70.5 & 67.5/\underline{90.2}  & \textbf{87.6}/-    & \underline{86.4}/\textbf{98.5} \\
Vase5       & 54.3/49.5 & 61.8/- & 52.0/- & 67.6/-    & -/47.2    & 54.8/61.5 & 85.2/\underline{93.7}  & \textbf{97.6}/-    & \underline{96.2}/\textbf{96.6} \\
Vase7       & 60.5/34.1 & 39.7/- & 46.2/- & 63.5/-    & -/70.1    & 51.3/75.8 & \textbf{96.6}/\underline{98.2}  & \underline{93.8}/-    & 89.5/\textbf{99.0}   \\
Vase8       & 48.5/50.7 & 52.9/- & 62.0/- & 63.0/-    & -/85.1    & 66.3/\underline{95.2} & \underline{73.9}/95.0  & 67.0/-    & \textbf{85.5}/\textbf{97.5} \\
Vase9       & 43.6/45.9 & 60.9/- & 59.4/- & 59.4/-    & -/61.5    & 70.3/74.4 & \textbf{83.0}/\textbf{95.2}  & 73.6/-    & \underline{81.5}/\underline{91.3} \\ \hline
Mean        & 51.1/54.9 & 55.9/- & 57.2/- & 66.1/65.0 & 75.7/69.1 & 61.9/74.5 & 83.9/\underline{89.8}  & \underline{84.2}/74.8 & \textbf{86.0}/\textbf{92.9}    \\ \bottomrule[1.5pt]
\end{tabular}}}
\end{table*}


\subsection{\textcolor{red}{F}. Detailed Results on MulSen-AD}~\label{Results_on_MulSen}

\begin{table*}[h]
\centering
\caption{\textbf{Quantitative Results on MulSen-AD.} The results are presented in O-ROC\%/P-ROC\%. The best performance is in \textbf{bold}, and the second best is \underline{underlined}.}
\label{table:mulsen}
\fontsize{11}{14}\selectfont{
\resizebox{\linewidth}{!}{
\begin{tabular}{c|cccccccc|
>{\columncolor{blue!8}}c
>{\columncolor{blue!8}}c
>{\columncolor{blue!8}}c
>{\columncolor{blue!8}}c
>{\columncolor{blue!8}}c
>{\columncolor{blue!8}}c}
\toprule[1.5pt]

Method~$\rightarrow$  & \small{M3DM-PM} & \small{M3DM-PB} & \small{PatchCore-FP} &\small{PatchCore-FP-R} &  \small{PatchCore-PM} &  \small{IMRNet}   & \small{Reg3D-AD} & \small{GLFM}   &\textbf{Simple3D}    \\
MulSen-AD & \small{CVPR'23}  & \small{CVPR'23} & \small{CVPR'22} & \small{CVPR'22}   & \small{CVPR'22} & \small{CVPR'24} & \small{NeurIPS'23} & \small{TASE'25}   & \textbf{Ours}   \\ \midrule

Capsule            & 73.1/77.7 & 67.1/75.3 & 89.8/91.7 & 90.5/91.9 & 90.3/92.1 & 60.1/42.3 & 91.2/87.7 & \underline{96.7}/\underline{93.0}  & \textbf{98.1}/\textbf{97.3} \\
Cotton             & 56.8/66.3 & \underline{80.5}/\textbf{69.9} & 25.3/55.4 & 26.3/54.6 & 19.7/52.8 & 58.5/50.7 & 43/52.1   & \textbf{81.2}/\underline{67.9}  & 78.8/60.2 \\
Cube               & 46.3/61.3 & 45.8/\underline{71.0}   & 72.3/57.5 & 66.8/43.7 & 72.2/41.7 & 43.2/56.6 & 56.9/62.6 & \underline{75.6}/68.0  & \textbf{82.7}/\textbf{73.2} \\
Spring pad         & 69.8/56.8 & 51.7/65.2 & 98.6/62.9 & \textbf{100}/60.1  & 96.5/62.1 & 65.1/40.1 & 95.1/\underline{80.2} & \textbf{100}/70.5 & \textbf{100}/\textbf{81.1}  \\
Screw              & 66.3/45.3 & 95.5/44.3 & \underline{97.9}/57.8 & 93.1/\underline{61.0}   & \textbf{99.7}/59.7 & 74.2/45.6 & 97.2/54.0   & 63.6/60.9  & 86.4/\textbf{74.7} \\
Screen             & 90.6/52.9 & \underline{92.8}/56.7 & 91.6/\textbf{60.9} & \textbf{95.0}/\underline{58.7}   & 89.7/53.2 & 37.8/35.2 & 64.1/46.6 & 86.6/50.8  & 73.8/52.9 \\
Piggy              & 16.4/61.7 & 44.7/57.2 & \textbf{100}/\underline{84.8}  & \underline{99.7}/62.4 & 98.2/60.3 & 72.9/51.2 & 86.6/63.5 & 73.2/77.8  & 89.6/\textbf{91.6} \\
Nut                & 78.3/63.1 & 75.1/68.7 & 97.1/90.3 & \textbf{98.9}/89.6 & \textbf{98.9}/89.7 & 81.2/36.9 & 79.7/80.7 & 94.0/\underline{95.7}  & 96.9/\textbf{97.2} \\
Flat pad           & 88.5/62.6 & 77.2/58.3 & \textbf{100}/70.7  & 89.3/67.8 & 94.4/63.0   & 71.4/54.2 & 90.8/69.2 & 94.6/\underline{77.2}  & \textbf{100}/\textbf{82.0}    \\
Plastic   cylinder & 46.2/51.0   & 70.6/65.2 & \underline{94.1}/\underline{83.0}   & 90.8/76.6 & 93.6/76.9 & 62.1/41.2 & 76.5/67.0   & 81.5/67.8  & \textbf{99.4}/\textbf{96.4} \\
Zipper             & 70.1/49.6 & 69.8/\underline{56.3} & 79.7/55.2 & \underline{81.3}/54.5 & 73.9/50.2 & 63.0/49.6   & 47.0/53.6   & \underline{81.3}/\textbf{57.4}  & \textbf{88.5}/53.0   \\
Button   cell      & 54.9/79.7 & 65.9/\underline{79.9} & \textbf{91.5}/38.2 & 68.7/51.2 & 79.7/47.8 & 70.2/48.5 & 78.2/70.6 & 56.7/43.3  & \underline{85.1}/\textbf{80.1} \\
Toothbrush         & 80.3/50.1 & \underline{90.1}/38.6 & \textbf{90.5}/60.5 & 88.8/60.4 & 89.1/\underline{60.6} & 61.5/51.9 & 81.2/47.2 & 84.9/57.8  & 82.9/\textbf{73.3} \\
Solar   panel      & 38.5/53.9 & 39.5/60.1 & 62.4/20.2 & 60.5/26.5 & 61.2/27.4 & 34.4/53.3 & \textbf{66.0}/\underline{60.9}   & 40.9/57.5  & \underline{64.4}/\textbf{94.9} \\
Light              & 57.9/48.0   & 65.3/49.5 & 97.5/\underline{70.7} & \textbf{100}/70.6  & \underline{99.2}/69.6 & 45.7/41.5 & 89.7/65.1 & 66.1/52.6  & 96.8/\textbf{95.9} \\  \hline
Mean               & 62.8/58.7 & 70.5/61.1 & \underline{86.0}/64.0     & 83.3/62.0   & 84/60.5   & 60.1/46.7 & 74.9/64.1 & 78.5/\underline{66.5}  & \textbf{88.2}/\textbf{80.3}  \\ \bottomrule[1.5pt]
\end{tabular}}}
\end{table*}


Quantitative comparison on the MulSen-AD dataset is shown in Tabel~\ref{table:mulsen}. Simple3D achieves the best performance with an average O-ROC of 88.2\% and P-ROC of 80.3\%, outperforming all existing baselines. Methods like GLFM, while effective in certain categories, fall short in generalization, with average P-ROC below 70\%. This reinforces the capability of Simple3D to generalize well across varying shapes, scales, and defect patterns, without reliance on complex augmentation or task-specific tuning.

\subsection{\textcolor{red}{G}. More Visualization of MiniShift}~\label{Visualization_MiniShift}

\begin{figure*}[t!]
\centering\includegraphics[width=\linewidth]{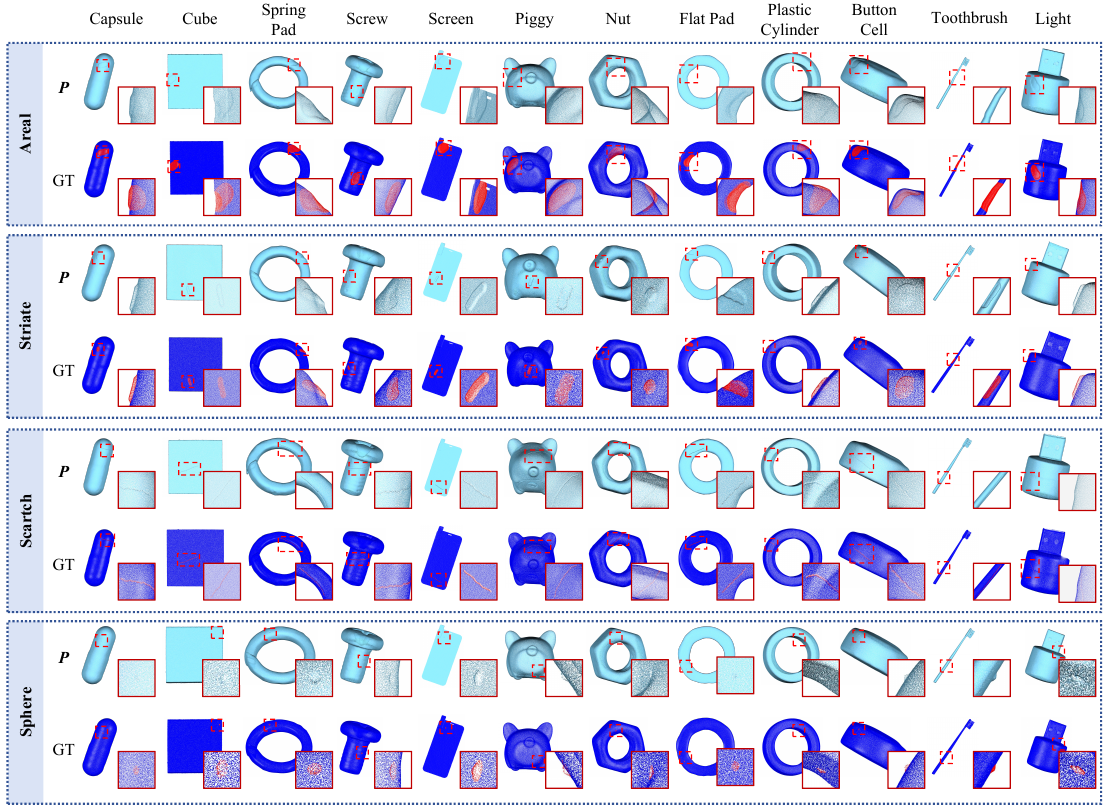}
\caption{\textbf{Visualization of data in MiniShift between four defect types and 12 categories.} The first row illustrates the high-resolution point clouds, and the second row presents their corresponding ground truth annotations. Defect regions are magnified to enable clearer visual observation.}
\vspace{-3mm}
\label{more_vis_minishift}
\end{figure*}

Figure~\ref{more_vis_minishift} illustrates representative samples from our proposed dataset \ourdataset{}, covering all four synthetic defect types (Areal, Striate, Scratch, and Sphere) across 12 object categories.  For each defect type, the top row presents high-resolution point cloud renderings of the defective instances, while the corresponding bottom row (GT) displays the ground truth annotations highlighting the exact defect regions.  Red boxes denote zoomed-in views that reveal localized structural anomalies in finer detail.  This visualization comprehensively demonstrates the diversity of object geometries and defect manifestations, validating the dataset’s capacity to support subtle 3D anomaly detection.

\subsection{\textcolor{red}{H}. More qualitative results}~\label{More_qualitative_results}

\begin{figure*}[t!]
\centering\includegraphics[width=\linewidth]{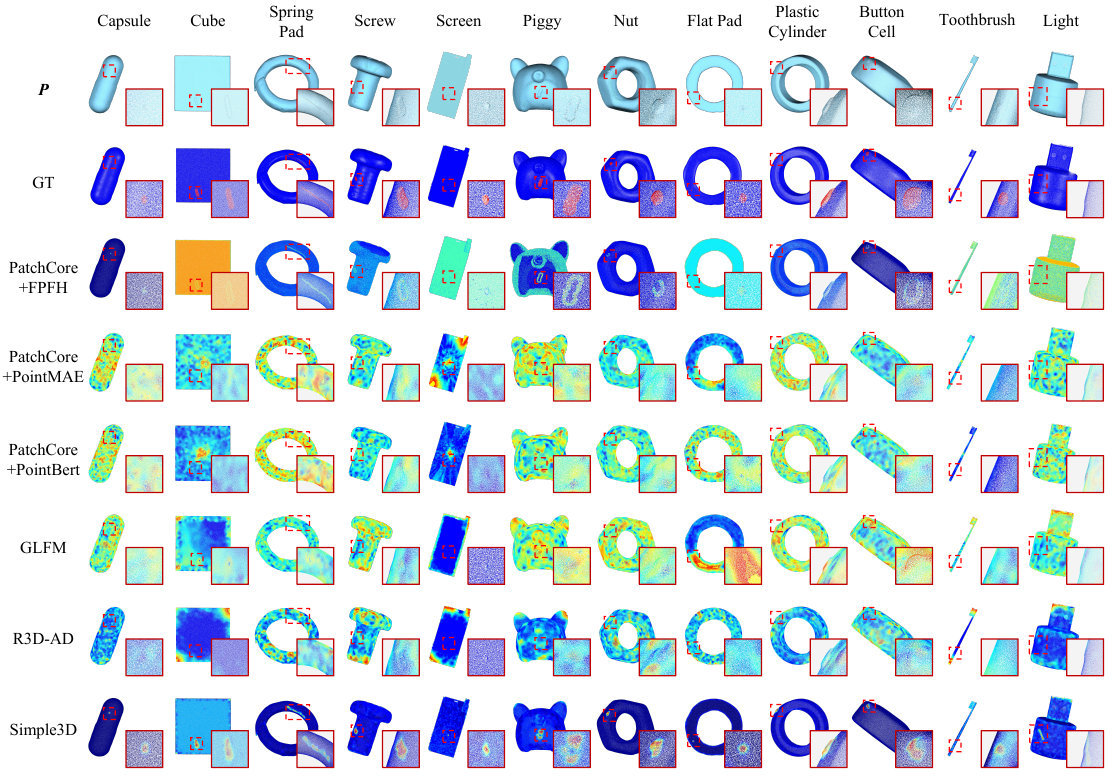}
\caption{\textbf{Visualization of prediction results in \ourdataset{} using the proposed method \ourmethod{} and other methods.} The first row is the original point clouds, while the second row is the ground truth. Subsequent rows depict various methods. Defect regions are magnified to enable clearer visual observation.}
\vspace{-3mm}
\label{more_qualitative_results}
\end{figure*}

Figure~\ref{more_qualitative_results} presents qualitative comparisons of point-wise anomaly detection results across \ourdataset{}. From top to bottom, we visualize the input point clouds, the ground-truth annotations (GT), and anomaly maps produced by competing methods, including PatchCore (including the features from FPFH, PointMAE, and PointBert), GLFM, R3D-AD, and our proposed Simple3D. Compared to existing methods, Simple3D consistently produces highly localized and sharply bounded anomaly maps, showing strong alignment with ground-truth defect regions. For example, on complex geometries such as the Screw, Plastic Cylinder, or Toothbrush, Simple3D effectively isolates subtle structural defects that are either missed or diffusely detected by other baselines. This visual evidence reinforces Simple3D’s ability to capture discriminative local descriptors, outperforming prior SOTA methods in both accuracy and interpretability across diverse defect scenarios.

\end{document}